\documentclass[11pt]{article}

\usepackage[preprint]{acl}

\usepackage{times}
\usepackage{latexsym}
\usepackage{amssymb}

\usepackage{booktabs}
\usepackage{multirow}
\usepackage{amssymb}
\usepackage{graphicx}
\usepackage[table]{xcolor}

\usepackage{amsmath}

\usepackage[T1]{fontenc}

\usepackage[utf8]{inputenc}

\usepackage{microtype}

\usepackage{inconsolata}

\usepackage{graphicx}

\title{LookStep: Efficient Vision-Language Navigation with Linguistic Foresight and Event Driven Memory}

\author{
 \textbf{Kun-Yang Yu\textsuperscript{1,2,3}\thanks{Work done as a remote intern in TermiTech}},
 \textbf{Yingzhe Li\textsuperscript{3}},
 \textbf{Hongyu Xu\textsuperscript{3}},
 \textbf{Shi-Yu Tian\textsuperscript{1,2}},
 \textbf{Zhi Zhou\textsuperscript{1,2 \dag}},
 \\
 \textbf{Yang Chen\textsuperscript{1,4}},
 \textbf{Ming Yang\textsuperscript{1,2}},
 \textbf{Sheng Wang\textsuperscript{3}},
 \textbf{Qing Yu\textsuperscript{3 \dag}},
 \textbf{Lan-Zhe Guo\textsuperscript{1,4}},
 \textbf{Yu-Feng Li\textsuperscript{1,2}\thanks{Corresponding author.}}
\\
 \textsuperscript{1}National Key Laboratory for Novel Software Technology, Nanjing University \\
 \textsuperscript{2}School of Artificial Intelligence, Nanjing University\\
 \textsuperscript{3}TermiTech\\
 \textsuperscript{4}School of Intelligence Science and Technology, Nanjing University\\
 \texttt{\{yuky,zhouz,liyf\}@lamda.nju.edu.cn, felix@termitech.cn}
}

\begin{document}
\maketitle
\begin{abstract}
Vision-Language Navigation (VLN) requires an embodied agent to follow natural-language instructions in unseen environments. Recent progress has been largely driven by Multimodal Large Language Models (MLLMs). Existing methods follow a next-step action prediction paradigm, supervising only the expert action, which requires a high quantity of data for training. They also rely on cognitive maps, accumulated historical frames, or external 3D tools to maintain states, leading to high computational and memory overhead. To realize resource efficiency VLN, we propose \textbf{LookStep}, a unified end-to-end framework that combines \textbf{Language Centric Future State Modeling} and \textbf{Event Driven Rolling Memory} that uses language labels to generate coarse-grained navigation progress and future states for each candidate action, while autonomously deciding whether to write each observation into a bounded rolling memory with a semantic role. We validate LookStep empirically. On VLN-CE tasks, LookStep outperforms existing methods under the same training settings, achieving a 49.7\% success rate on R2R-CE Val-Unseen with better memory efficiency and less data usage. Code and model is available at \url{https://github.com/kunyang-YU/LookStep}.
\end{abstract}

\section{Introduction}
Vision-Language Navigation (VLN) is a fundamental task in embodied intelligence, requiring an agent to perform navigation in real or simulated environments according to natural-language instructions and visual observations. In recent years, with the rapid development of Multimodal Large Language Models (MLLMs)~\citep{lan2025mappo,lan2026contextual,qi2026mernav}, an increasing number of studies have begun to use MLLMs as end-to-end navigation models, leveraging their visual perception, semantic understanding, and language reasoning abilities to directly predict navigation actions from instructions and visual inputs~\citep{zhang2024uni,liang2026persistent,yuan2026unimapgen,shan2025stability}.

Existing MLLM-based VLN methods usually introduce additional state modeling mechanisms beyond action prediction. One line of work converts visual scenes into textual descriptions and continuously maintains semantic cognitive maps~\citep{zhang2025mapnav,zeng2024driving,chen2024spatialvlm,liu2025navforesee} or navigation memory in language form. Another line of work directly preserves historical visual frames and feeds them together with the current observation into the model for decision-making~\citep{cheng2024navila,yang2025nav,xiang2025nav,wei2025streamvln}. In addition, some methods introduce external visual-spatial tools or geometric modeling modules to compensate for the limited 3D spatial understanding of general-purpose visual encoders~\citep{zeng2025janusvln}.

\begin{figure}
    \centering
    \includegraphics[width=\linewidth]{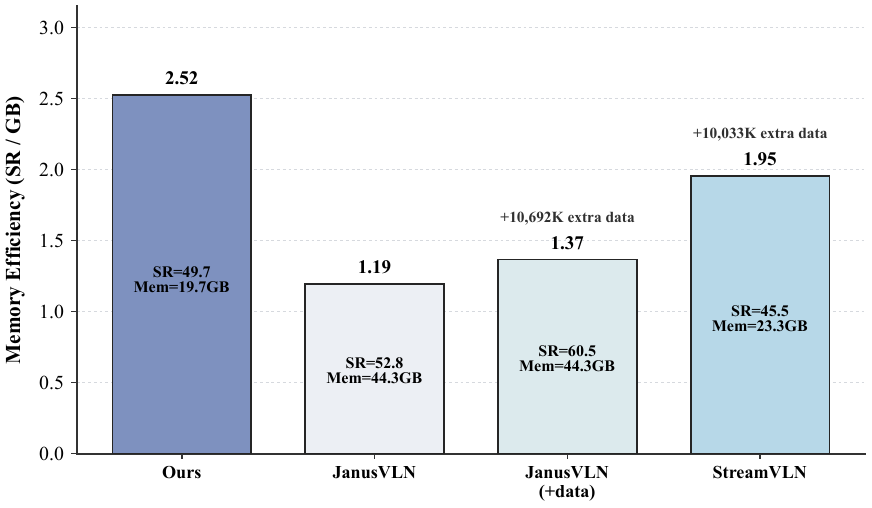}
    \caption{Memory efficiency comparison of different methods in R2R. LookStep performs the best memory efficiency and does not require extra data for training.}
    \label{fig:aa}
    \vspace{-1.35em}
\end{figure}
Although these methods improve navigation state awareness, they still suffer from clear limitations. Textual scene descriptions can hardly preserve spatial relations, directional information, and fine-grained visual changes in a complete manner, which requires a large quantity of data for training. Directly accumulating historical observations introduces substantial computational overhead, while fixed-window, uniform sampling, or heuristic sampling strategies may retain redundant frames and miss key navigation events. Relying on external spatial tools can provide additional geometric information, but it also significantly increases system complexity and GPU memory requirements, weakening the efficiency for practical deployment.

Therefore, efficient VLN is not merely a problem of state representation, but relies on the synergy between efficient data utilization and memory management. In this end, we propose \textbf{LookStep}, an VLN framework for continuous VLN, consisting of two modules: \textbf{Language Centric Future State Modeling} and \textbf{Event Driven Rolling Memory}. Language Centric Future State Modeling reformulates action prediction from direct next-step into future state modeling action evaluation to better use the data in training. 
Viewed through the lens of test-time learning~\cite{DBLP:conf/nips/ZhouYLYGLM25,DBLP:conf/aaai/ZhouYGL25,DBLP:conf/emnlp/TianZYYJGL25,DBLP:conf/acl/TianZDYYCGL26}, Event Driven Rolling Memory can be regarded as a form of online adaptation in the memory space that requires no parameter updates. The model actively maintains navigation-critical historical information by autonomously determining, at each step, whether the current observation should be written into a bounded episodic memory and what semantic role it should play, thereby continually updating the agent’s episodic state.


Furthermore, we provide an oracle-level information-theoretic motivation showing that expert-derived future-state labels can contain action-relevant information. Empirically, we evaluate the effectiveness and efficiency of LookStep on two VLN-CE benchmarks. As shown in Figure~\ref{fig:aa},  
LookStep enables data- and memory-efficient navigation without relying on large-scale additional data or auxiliary spatial modeling tools, and achieves SOTA performance among methods under the same training settings, reaching a success rate of 49.7\% on R2R-CE Val-Unseen.

Our contributions are summarized as follows:

\begin{itemize}
    \item We propose \textbf{LookStep}, a resource efficiency VLN framework. The framework explicitly models navigation progress and candidate-action future states before predicting the final action, and actively preserves navigation-critical historical observations through model-generated memory write decisions and semantic memory roles.

    \item We provide an oracle-level information-theoretic motivation showing that expert-derived future-state labels can contain action-relevant information.

    \item We conduct extensive experiments on the VLN-CE benchmarks to validate the effectiveness and efficiency of our method. The results show that, using only monocular RGB observations and without relying on external spatial modeling tools, our framework still achieves competitive navigation performance with data and memory efficiency.
\end{itemize}
\section{Related Work}

\subsection{Vision-Language Navigation}

Vision-Language Navigation (VLN)~\citep{krantz2020beyond,lu2024scaling}, an important embodied tasks~\cite{DBLP:journals/corr/abs-2605-01293,chen2025re}, requires an embodied agent to follow natural-language instructions and navigate to target locations from visual observations. Early VLN methods~\citep{zheng2025towards,hong2022bridging,du2024self} formulate navigation as discrete decision making over panoramic viewpoints \cite{DBLP:conf/cvpr/AndersonWTB0S0G18}, while continuous VLN further requires low-level action execution in realistic simulators \cite{DBLP:journals/corr/abs-2004-02857, DBLP:conf/emnlp/KuAPIB20}. Video-based MLLMs and vision-language-action models make decisions through imitation learning or neuro-symbolic reasoning~\cite{DBLP:conf/ijcai/YangSGZZJDL25}, such as NaVid~\citep{zhang2024navid}, Uni-NaVid~\citep{zhang2024uni}, and NaVILA~\citep{cheng2024navila}, have been adopted to predict actions directly from egocentric RGB observations.

Recent studies have improved the performance of MLLM-based VLN methods. StreamVLN adopts slow-fast context modeling for streaming action generation with bounded context~\cite{wei2025streamvln}. JanusVLN introduces dual implicit memories to enhance RGB-only spatial reasoning and reduce redundant history computation~\cite{zeng2025janusvln}. However, they still rely on large-scale data under the next-step prediction paradigm and imitation learning~\cite{DBLP:journals/corr/abs-2411-18201}. We propose \textbf{Language Centric Future State Modeling}, an auxiliary task that improves data utilization and strengthens navigation progress understanding.

\subsection{Spatial Reasoning via MLLMs}

Spatial reasoning is crucial for VLN. Recent works enhance the ability of spatial reasoning in MLLMs.~\citep{zeng2026futuresightdrive,tian2026self,DBLP:journals/corr/abs-2604-09712,DBLP:journals/corr/abs-2603-16307,DBLP:journals/corr/abs-2603-24004} SpatialVLM uses large-scale spatial VQA data \cite{chen2024spatialvlm}. SpatialRGPT performs region-level reasoning with depth-enhanced representations \cite{cheng2024spatialrgpt}, and 3D-LLM, LEO, and Scene-LLM inject 3D scene representations into LLMs for embodied reasoning and planning \cite{hong20233d, huang2023embodied, fu2024scene}. Moreover, many world-modeling approaches, such as JEPA~\cite{DBLP:journals/corr/abs-2306-02572} and DreamerV3~\cite{DBLP:journals/corr/abs-2301-04104}, learn latent dynamics, predict future representations, or model transitions between environmental states to reason about future outcomes during task execution and process memory representations. These methods have achieved remarkable success in their respective application domains~\cite{DBLP:journals/corr/abs-2603-24257,DBLP:conf/icra/LiuGWCPSP25,DBLP:conf/iclr/ParkCA25}.

Another line of work improves spatial reasoning from RGB or video inputs~\citep{li2024mlp,liu2026fedadamw,yang2025fast3r}. Spatial-MLLM incorporates geometry-aware visual features for 2D-based spatial intelligence \cite{wu2026spatial}. VGGT provides feed-forward 3D geometry priors for downstream spatial representation learning \cite{wang2025vggt}. We propose \textbf{Event Driven Rolling Memory}, which leverages the spatial reasoning ability of MLLMs to preserve key navigation events, enabling efficient navigation.

\section{Method and Theoretical Insights}

\subsection{Navigation Task Definition}

The Vision-and-Language Navigation task in continuous environments is defined as follows. At timestamp $t$, an embodied agent is provided with a natural language instruction $I$ and an ego-centric RGB observation $x_t\in \mathrm{R}^{3\times H\times W}$. The observation sequence is $O_t = \{x_0,\dots,x_t\}$ and the action space is $A=\{\texttt{MOVE},\texttt{TURN\_LEFT},\texttt{TURN\_RIGHT},\texttt{STOP}\}$. The goal is to predict the next action $a_{t+1} \in A$. Each action corresponds to a fine-grained physical change. After the action is executed, a new observation $x_{t+1}$ is obtained. This process iterates until the agent executes the \textbf{Stop} action at the target location as specified by the instruction.

\subsection{Methods Architecture}

\begin{figure*}
    \centering
    \includegraphics[width=\linewidth]{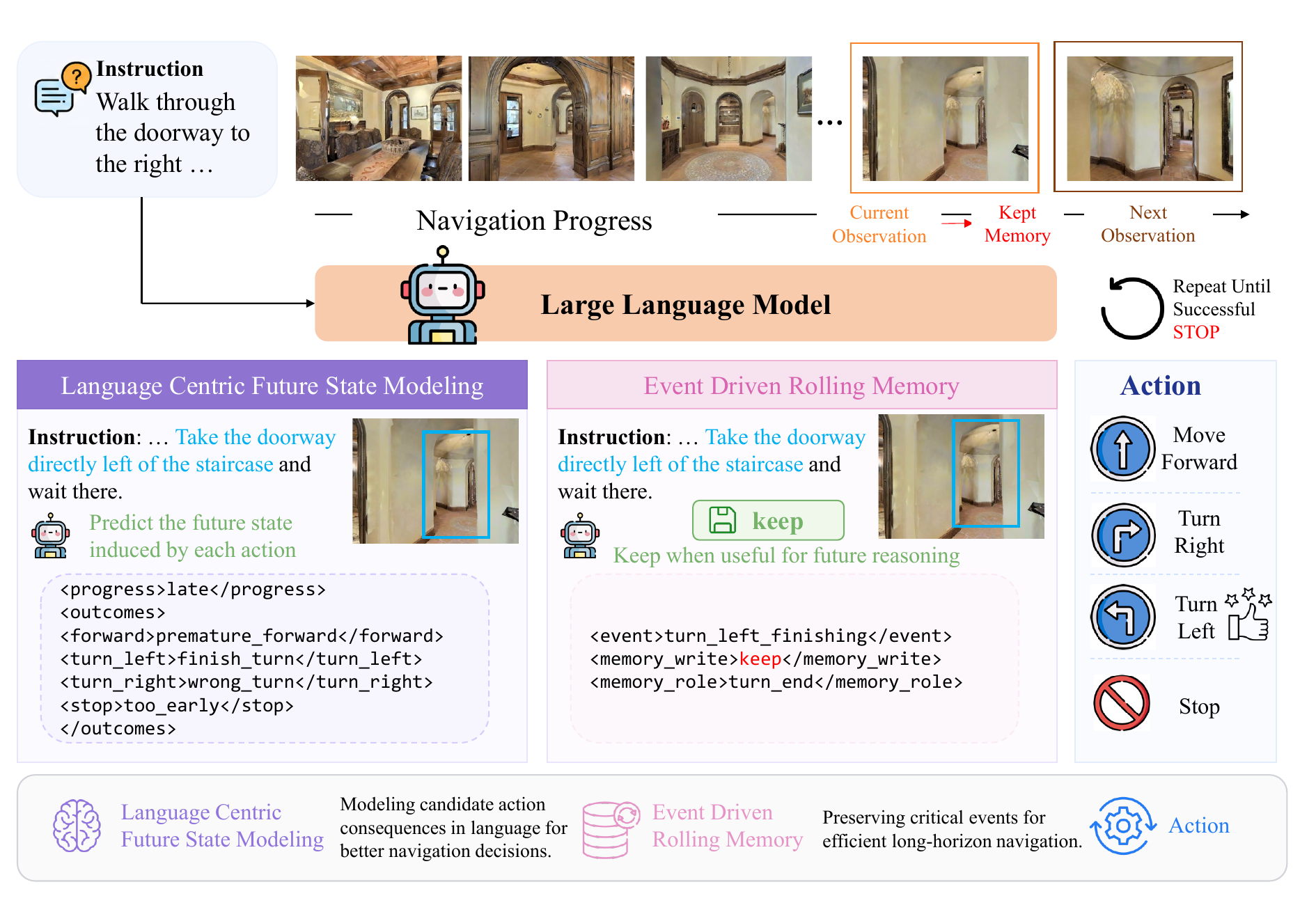}
    \caption{Structure of LookStep approach. It uses language tags in Language Centric Future State Modeling and Event Driven Rolling Memory to realize data and memory efficiency navigation.}
    \label{fig:structure}
\end{figure*}

Existing VLN methods typically predict the next action directly from language instructions, often with the assistance of visual-side spatial modules. Current approaches suffer from two orthogonal limitations in both data usage and memory management. On the one hand, the prevailing next-step action prediction paradigm requires large-scale expert data supervision, limiting the data efficiency and scalability of VLN training. Moreover, its dependence on external spatial modeling modules increases memory cost and underutilizes the intrinsic visual-spatial understanding and language reasoning abilities of MLLMs. On the other hand, existing memory management strategies mainly rely on memory sampling, which may fail to preserve navigation-relevant historical information and therefore introduce ineffective or missing memory contexts during long-horizon navigation.

To address the above two limitations, we propose \textbf{LookStep}, whose overall framework is illustrated in Figure~\ref{fig:structure}. LookStep consists of two complementary components: \textbf{Language Centric Future State Modeling} and \textbf{Event Driven Rolling Memory}.
First, to overcome the limitations of existing VLN methods that mainly rely on next-step action imitation and suffer from inefficient data utilization, \textbf{Language Centric Future State Modeling} leverages the visual understanding and linguistic expression abilities of MLLMs to represent the short-term navigation consequence of each candidate action as a compact language label. Given the VLN input, the model jointly predicts the current coarse-grained navigation progress, the future navigation state associated with each candidate action, and the final action. In this way, VLN training is transformed from simple action imitation into future-oriented action evaluation. This design enables the model to make fuller use of trajectory data during training and enhances its understanding of actions during navigation.
Second, to address the redundancy of historical observations and the difficulty of preserving key events, \textbf{Event Driven Rolling Memory} models historical context as a memory bank dynamically updated by navigation events. At each step, the model not only predicts the next action, but also determines whether the current observation should be written into long-term memory and assigns it a semantic role. During inference, key events trigger memory writes. When the memory exceeds its capacity, the earliest entries are removed in a first-in-first-out (FIFO) manner. This strategy reduces computational overhead while retaining the most critical historical information for navigation decision-making.

Formally, given the instruction $I$, the current observation $x_t$, and the historical observations $O_{t-1}$, the model is required to predict three components in natural language:

\begin{equation}
    \text{MLLM} (I, O_{t-1},x_t)  =  \{\mathcal{P}_t,\ \mathcal{M}_t,\ a_{t+1}\}
\end{equation}

where $\mathcal{P}_t$ is the future state prediction and $\mathcal{M}_t$ is the memory management state. We will introduce them in detail below.

\subsection{Language Centric Future State Modeling}

We adopt a \textbf{L}anguage Centric \textbf{F}uture \textbf{S}tate Modeling (LFS) strategy to explicitly model the possible navigation outcomes of candidate actions, which enables the model to make fuller use of trajectory data during training and enhances its understanding of actions during navigation.

Specifically, $\mathcal{P}_t$ consists of two aspects of the navigation process. Firstly, we predict the current coarse-grained navigation progress $p_t$, which strengthens the model's understanding of its current position, path completion status, and proximity to the target. We use a predefined label set to describe the navigation progress. Secondly, we model the short-term navigation outcome $F_t$ that each candidate action in the action space may lead to, enabling the model to compare the consequences of different actions before producing the final action. For each action in the action space, we use an outcome label set to describe the possible outcomes.

\begin{equation}
    \mathcal{P}_t = \{ p_t, F_t \}
\end{equation}

For effective training and prediction, we organize the corresponding outcomes into a structured language sequence using the tags described below. 

\begin{itemize}
    \item \texttt{<Progress>}$p_t$\texttt{</Progress>}: current coarse grained navigation progress
    \item \texttt{<Outcome>}$F_t$\texttt{</Outcome>}:future navigation states outcomes
\end{itemize}

All outcomes are selected from a predefined label set designed according to navigation-specific rules to ensure output consistency and controllability. Detailed definitions of these labels are provided in the Appendix~\ref{append:labelset}.

\subsection{Event Driven Rolling Memory}

We adopt \textbf{E}vent \textbf{D}riven \textbf{R}olling \textbf{M}emory (EDRM) to address the loss of critical historical information and contextual inconsistency caused by sampling-based history modeling methods. Our method introduces an active memory selection mechanism, where the navigation model autonomously determines whether the current observation constitutes a navigation-critical event, whether it should be incorporated into historical memory, and what semantic role it plays in the navigation process.

At each navigation step, the model performs unified memory management, denoted as $\mathcal{M}_t$. The goal of $\mathcal{M}_t$ is to actively identify navigation-critical information, such as key action transitions, important landmarks, and goal-related evidence, and preserve them as part of the historical context. This enhances the informativeness of historical observations for estimating the current navigation state. Specifically, $\mathcal{M}_t$ consists of two components: a memory write label $W_t$, which determines whether the current observation should be stored in historical memory, and a memory role label $R_t^{m}$, which specifies the semantic role of this observation in the navigation process:

\begin{equation}
    \mathcal{M}_t = \{W_t, R_t\}
\end{equation}

Similar to the Language Centric Future State Modeling, the memory management state is also organized in language sequence, with a detailed predefined label set described in the Appendix~\ref{append:labelset}:

\begin{itemize}
    \item \texttt{<MemoryWrite>}$W_t$\texttt{</MemoryWrite>}: indicates whether the current observation should be written into long-term memory.
    \item \texttt{<MemoryRole>}$R_t$\texttt{</MemoryRole>}: describes the semantic role of the current observation in the navigation process.
\end{itemize}

During inference, the model updates the event memory online according to the generated memory write label $W_t$. When $W_t=\text{keep}$, the current observation $x_t$ and its corresponding memory role $R_t^{m}$ are written into memory. When $W_t=\text{drop}$, the current observation is not added to memory and is only temporarily retained as part of the recent observation context. We implement the memory with first-in-first-out (FIFO) queues, which enable lightweight rolling memory updates while preserving the most critical historical information for navigation decision-making.

\subsection{Training Process}
We train LookStep to autoregressively generate, at every step $t$ along an expert trajectory, a single structured target that unifies the two components introduced above:
\begin{equation}
\mathcal{T}_t^{\star}=\bigl(p_t^{\star},\,F_t^{\star},\,W_t^{\star},\,R_t^{\star},\,a_{t+1}^{\star}\bigr).
\end{equation}
The action $a_{t+1}^{\star}$ is taken directly from the ground-truth action label, while the progress $p_t^{\star}$, candidate-action outcomes $F_t^{\star}$, memory-write decision $W_t^{\star}$, and memory role $R_t^{m,\star}$ are derived automatically from the ground-truth trajectory according to the labelling rules detailed in the Appendix~\ref{append:labelset}. Denoting the input context as $X_t=(I, O_{t-1},x_t)$, the model factorizes the joint distribution autoregressively in order. Let $w_1^{\star},\dots,w_{K_t}^{\star}$ denote the tokenization of $\mathcal{T}_t^{\star}$. The training objective is the standard next-token negative log-likelihood over the full structured sequence:
\begin{equation}
\mathcal{L}(\theta)
=-\,\mathbb{E}_{(X_t,\mathcal{T}_t^{\star})\sim\mathcal{D}}
   \sum_{k=1}^{K_t}\log P_\theta\!\left(w_k^{\star}\,\middle|\,X_t,\,w_{<k}^{\star}\right).
\label{eq:training-loss}
\end{equation}
Unlike conventional VLN training that only back-propagates through the action token, Eq.~\eqref{eq:training-loss} simultaneously supervises progress estimation, future-state imagination, memory-event recognition, and final action selection under a single loss. As a result, the final action prediction is compelled to route through the language-form future-state variable $F_t$ rather than being inferred directly from $X_t$, converting otherwise implicit reasoning over candidate-action consequences into an explicit and supervised intermediate signal. Meanwhile, key-event recognition for memory management is internalized into the same autoregressive generation process without any auxiliary objective or hand-crafted heuristics. More importantly, this auxiliary task design extends supervision from action-level imitation to trajectory-level reasoning, thereby improving data utilization. The label of LookStep used in the training process is constructed based on the expert trajectory, which is detailed described in the Appendix~\ref{append:labelset}

\begin{table*}[ht]
\centering
\caption{
Comparison with SOTA methods on VLN-CE R2R and RxR Val-Unseen splits. 
External data includes any sources beyond the standard R2R/RxR-CE datasets.
All results are from their respective papers. 
Pano, Odo, Depth, and S.RGB respectively represent panoramic view, odometry, depth, and single RGB. NaVILA* excludes human-following data. JanusVLN means 0K extra training data version. StreamVLN* uses EnvDrop as external data version.
}
\label{tab:vlnce_r2r_rxr_val_unseen}
\resizebox{\textwidth}{!}{
\begin{tabular}{l|cccc|cc|cccc|ccc}
\toprule
\multirow{2}{*}{Method} 
& \multicolumn{4}{c|}{Observation} 
& \multicolumn{2}{c|}{Training}
& \multicolumn{4}{c|}{R2R Val-Unseen}
& \multicolumn{3}{c}{RxR Val-Unseen} \\
\cmidrule(lr){2-5} \cmidrule(lr){6-7} \cmidrule(lr){8-11} \cmidrule(lr){12-14}
& Pano. & Odo. & Depth & S.RGB 
& Auxiliary Modules & External Data
& NE$\downarrow$ & OS$\uparrow$ & SR$\uparrow$ & SPL$\uparrow$
& NE$\downarrow$ & SR$\uparrow$ & SPL$\uparrow$ \\
\midrule

HPN+DN 
& $\checkmark$ & $\checkmark$ & $\checkmark$ & 
& - & -
& 6.31 & 40.0 & 36.0 & 34.0
& - & - & - \\

CMA 
& $\checkmark$ & $\checkmark$ & $\checkmark$ & 
& - & -
& 6.20 & 52.0 & 41.0 & 36.0
& 8.76 & 26.5 & 22.1 \\

Sim2Sim 
& $\checkmark$ & $\checkmark$ & $\checkmark$ & 
& - & -
& 6.07 & 52.0 & 43.0 & 36.0
& - & - & - \\

VLN$\circlearrowright$BERT 
& $\checkmark$ & $\checkmark$ & $\checkmark$ & 
& - & -
& 5.74 & 53.0 & 44.0 & 39.0
& 8.98 & 27.0 & 22.6 \\

Ego$^2$-Map 
& $\checkmark$ & $\checkmark$ & $\checkmark$ & 
& - & -
& 5.54 & 56.0 & 47.0 & 41.0
& - & - & - \\

DreamWalker 
& $\checkmark$ & $\checkmark$ & $\checkmark$ & 
& - & -
& 5.53 & 59.0 & 49.0 & 44.0
& - & - & - \\

GridMM 
& $\checkmark$ & $\checkmark$ & $\checkmark$ & 
& - & -
& 5.11 & 61.0 & 49.0 & 41.0
& - & - & - \\

Reborn 
& $\checkmark$ & $\checkmark$ & $\checkmark$ & 
& - & -
& 5.40 & 57.0 & 50.0 & 46.0
& 5.98 & 48.6 & 42.0 \\

InstructNav 
& $\checkmark$ & $\checkmark$ & $\checkmark$ & 
& - & -
& 6.89 & - & 31.0 & 24.0
& - & - & - \\

\midrule

COSMO 
& $\checkmark$ &  &  & 
& - & -
& - & 56.0 & 47.0 & 40.0
& - & - & - \\

AO-Planner 
& $\checkmark$ &  & $\checkmark$ & 
& - & -
& 5.55 & 59.0 & 47.0 & 33.0
& 7.06 & 43.3 & 30.5 \\

LAW 
&  & $\checkmark$ & $\checkmark$ & $\checkmark$
& - & -
& 6.83 & 44.0 & 35.0 & 31.0
& 10.90 & 8.0 & 8.0 \\

MapNav 
&  & $\checkmark$ & $\checkmark$ & $\checkmark$
& - & -
& 4.93 & 53.0 & 39.7 & 37.2
& - & - & - \\

g3D-LF 
&  & $\checkmark$ & $\checkmark$ & $\checkmark$
& - & -
& 5.70 & 59.5 & 47.2 & 34.6
& - & - & - \\

Seq2Seq 
&  &  & $\checkmark$ & $\checkmark$
& - & -
& 7.77 & 37.0 & 25.0 & 22.0
& 12.10 & 13.9 & 11.9 \\

Navid-4D 
&  &  & $\checkmark$ & $\checkmark$
& - & -
& 5.99 & 55.7 & 43.8 & 37.1
& - & - & - \\

NavMorph 
&  &  & $\checkmark$ & $\checkmark$
& - & -
& 5.75 & 56.9 & 47.9 & 33.2
& 8.85 & 30.8 & 22.8 \\

\midrule

NaVid 
&  &  &  & $\checkmark$
& $\checkmark$ & 953K
& 5.47 & 49.1 & 37.4 & 35.9
& - & - & - \\

Uni-NaVid 
&  &  &  & $\checkmark$
& $\checkmark$ & 3577K
& 5.58 & 53.3 & 47.0 & 42.7
& 6.24 & 48.7 & 40.9 \\

NaVILA* 
&  &  &  & $\checkmark$
& $\checkmark$ & 12574K
& 5.37 & 57.6 & 49.7 & 45.5
& 6.77 & 49.3 & 44.0 \\

JanusVLN
&  &  &  & $\checkmark$
& $\checkmark$ & 0K
& 5.17 & 58.0 & 52.8 & 49.2
& 6.46 & 51.4 & 44.3 \\

\midrule

Sim2Real 
&  &  &  & $\checkmark$
&  & 0K
& 5.95 & 55.8 & 44.9 & 30.4
& 8.79 & 36.7 & 25.5 \\

StreamVLN* 
&  &  &  & $\checkmark$
&  & 10033K
& 6.05 & 53.8 & 45.5 & 41.6
& - & - & - \\

\rowcolor{gray!20}
\textbf{Ours}
&  &  &  & $\checkmark$
&  & 0K
& \textbf{5.34} & \textbf{55.9} & \textbf{49.7} & \textbf{45.3}
& \textbf{6.89} & \textbf{46.9} & \textbf{39.9} \\

\bottomrule
\end{tabular}
}
\end{table*}

\subsection{Theoretical Motivation}
  \label{sec:theory}

  We provide an oracle-level information-theoretic perspective to motivate the design of Language-Centric Future State Modeling. Let $X_t$ denote the navigation context, $a_{t+1}^{\star}$ the expert action, and $F_t^{\star}$ the action-relevant future-
  state labels constructed from the expert trajectory. Under log loss, the Bayes-optimal risks of direct action prediction and oracle future-state-conditioned prediction are
  \begin{equation}
      \mathcal{R}_{\mathrm{base}}^{\star}
      = H(a_{t+1}^{\star}\mid X_t),
\end{equation}
\begin{equation}
      \mathcal{R}_{\mathrm{oracle\text{-}fsm}}^{\star}
      = H(a_{t+1}^{\star}\mid X_t,F_t^{\star}).
  \end{equation}
  Their difference is
  \begin{equation}
  \begin{aligned}
      \mathcal{R}_{\mathrm{base}}^{\star}
      -\mathcal{R}_{\mathrm{oracle\text{-}fsm}}^{\star}
      &=
      I(a_{t+1}^{\star};F_t^{\star}\mid X_t)
      \geq 0.
  \end{aligned}
  \end{equation}
  The inequality is strict when
  $I(a_{t+1}^{\star};F_t^{\star}\mid X_t)>0$.
  This identity shows that an oracle future-state representation can reduce action uncertainty by explicitly exposing information relevant to candidate-action selection. It therefore motivates our design of structured candidate-action outcomes. Additionally, at inference time, LookStep does not observe $F_t^{\star}$, but generates $\hat{F}_t$, this may introduce the gap between optimal predictor and real situation. We provide a detailed analysis about this gap in Appendix~\ref{app:theory111} and our experiment result in Section 4 shows the effectiveness of the Language Centric Future State Modeling.

\begin{figure*}[htbp]
    \centering
    \includegraphics[width=\linewidth]{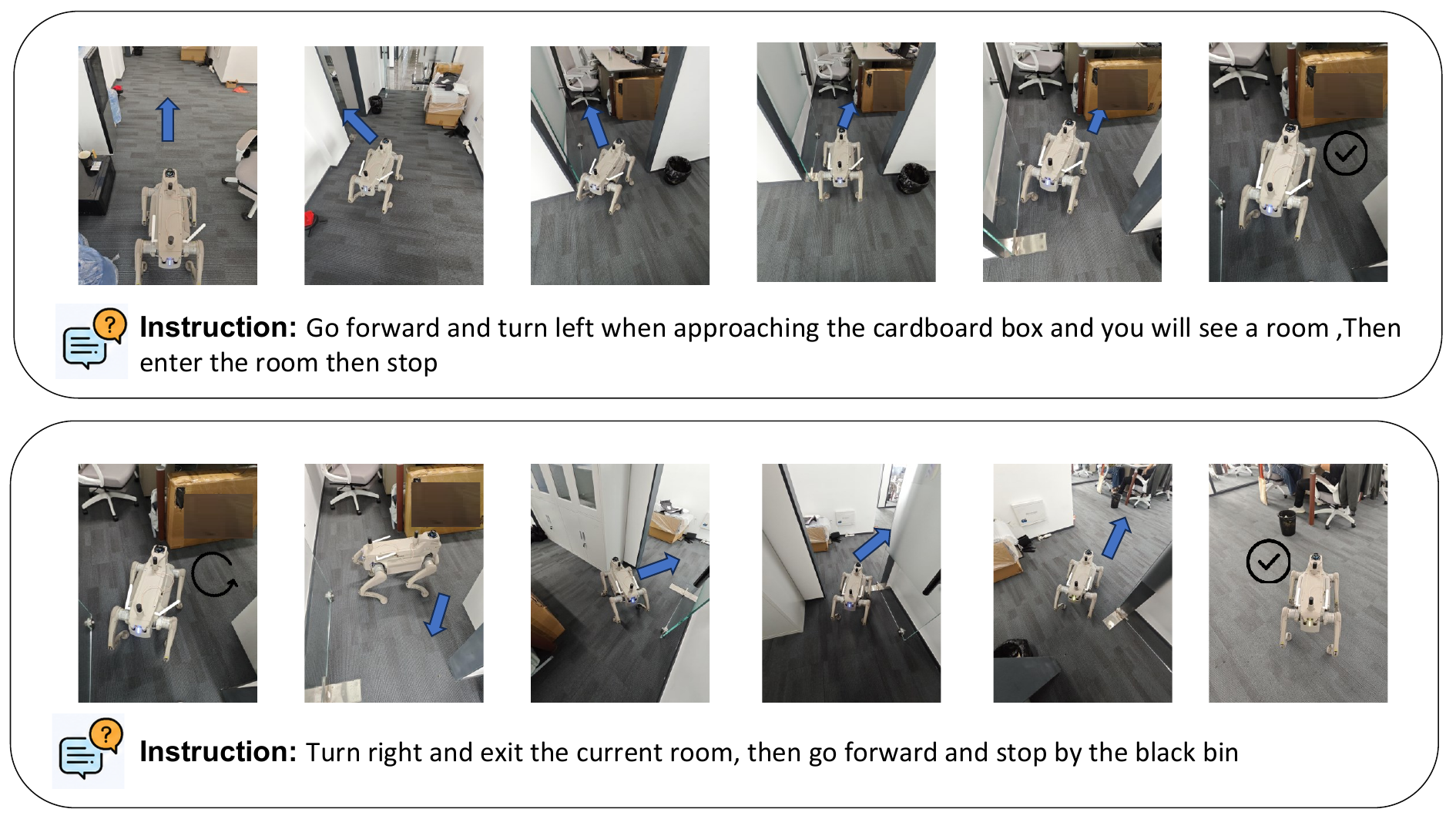}
    \caption{Qualitative results of our methods on real-world.}
    \label{fig:my_label}
\end{figure*}

\section{Experiments}
\subsection{Experiment Setup}
\paragraph{Simulation environments and metrics.} 

We conduct our experiments on  two of the most recognized benchmark datasets~\citep{krantz2020beyond}: R2R-CE~\citep{DBLP:conf/cvpr/AndersonWTB0S0G18} and RxR-CE~\citep{DBLP:conf/emnlp/KuAPIB20}. These datasets comprise trajectories collected from Matterport3D~\citep{chang2017matterport3d} and using the Habitat simulator~\citep{savva2019habitat}. We report performance on the unseen splits using standard VLN metrics, including Navigation Error (NE), Oracle Success Rate (OS), Success Rate (SR), and Success-weighted Path Length (SPL).

\paragraph{Implementation details.}

We constructed our methods based on Qwen3-VL 8B~\citep{bai2025qwen3}. The model is trained for one epoch using only the R2R-CE and RxR-CE datasets. We set the memory length to 8 for the memory queue. We provide a detailed implementation setting in the Appendix~\ref{append:labelset}.

\subsection{Main Results}

\paragraph{Results on VLN-CE benchmark. }
As shown in Table~\ref{tab:vlnce_r2r_rxr_val_unseen}, we systematically evaluate our method on two VLN-CE datasets. Although our method uses only a single RGB image as input, without relying on panoramic observations, odometry, or other additional sensor signals, it still achieves about a 5\% improvement in success rate over methods using richer input modalities, with gains reaching up to 20\% in some settings. This indicates that performance improvements in VLN do not necessarily require complex input forms, but can also come from a more effective navigation paradigm.

Furthermore, compared with methods following the conventional \textit{next-step prediction} paradigm, our method also achieves consistent advantages. Whether these methods use explicit textual cognitive maps, such as MapNav, or historical frames as context, such as StreamVLN, our method brings about a 5\% performance improvement in R2R-CE dataset. This suggests that simply enhancing historical context or constructing intermediate map representations is still insufficient for long-horizon navigation decision-making. In contrast, our method more effectively models candidate actions and their consequences, thereby improving the quality of action selection, which is also consistent with our theoretical analysis.

More importantly, our method is trained only on offline expert trajectories and does not rely on additional data collection strategies~\citep{tan2019learning,wang2023scaling} such as DAgger~\citep{ross2011reduction}, yet it still outperforms existing methods such as NaVid and StreamVLN. This shows that the performance gain of our method does not come from additional interaction data, but from a more effective navigation modeling paradigm. Moreover, compared with the current state-of-the-art method JanusVLN, our method achieves competitive performance. It is worth emphasizing that JanusVLN relies on external visual spatial modeling tools such as VGGT, whereas our method requires no additional visual tools. Therefore, while approaching state-of-the-art performance, our method significantly reduces system complexity and computational overhead. We further quantify this advantage in the \textit{Memory Length and Efficiency} section.

\paragraph{Real-World Experiments}
We conduct real-world experiments in a challenging workspace environment. This scene contains multiple visually similar and easily confusable objects and involves complex navigation instructions such as entering and exiting rooms. As shown in the Figure 3, LookStep successfully completes difficult navigation tasks in such complex scenarios. Notably, LookStep is trained entirely on simulated data, and these results therefore demonstrate its strong generalization ability to novel tasks and unseen scenes.
Details are described in Appendix~\ref{append:labelset}
\subsection{Further Analysis}

\paragraph{Ablation Study}

\begin{table}[t]
\centering
\caption{Ablation study of our modules in R2R dataset. * denotes without re-training the model}
\label{tab:ablation_memory}
\begin{tabular}{l|cccc}
\toprule
Method & NE$\downarrow$ & OS$\uparrow$ & SR$\uparrow$ & SPL$\uparrow$ \\
\midrule
Ours & \textbf{5.34} & 55.9 & \textbf{49.7} & \textbf{45.3} \\
Ours w/o LFS & 5.39 & 52.8 & 46.9 & 42.4 \\
Ours w/o EDRM* & 6.64 & \textbf{70.9} & 37.4 & 22.1 \\
\bottomrule
\end{tabular}
\end{table}
We conduct ablation studies to verify the effectiveness of each component in LookStep. As shown in Table~\ref{tab:ablation_memory}, removing any component leads to a drop in success rate, indicating that both LFS and EDRM are indispensable. Removing LFS degrades all metrics, demonstrating that language-centric future state modeling effectively enhances candidate-action consequence reasoning, which is consistent with our theoretical analysis. For EDRM, we replace the event-driven memory with the uniform history sampling strategy used in JanusVLN during inference, which results in a clear decrease in success rate. This shows that event-level historical memory is crucial for long-horizon navigation. Notably, OS instead improves after removing EDRM, suggesting that the model can still approach the target region with the help of LFS, but struggles to accurately determine when to stop without key historical information. Therefore, LFS mainly helps the model select the correct path, while EDRM primarily supports target confirmation and stopping decisions.

\paragraph{Memory Length and Efficiency}

\begin{figure}
    \centering
    \includegraphics[width=\linewidth]{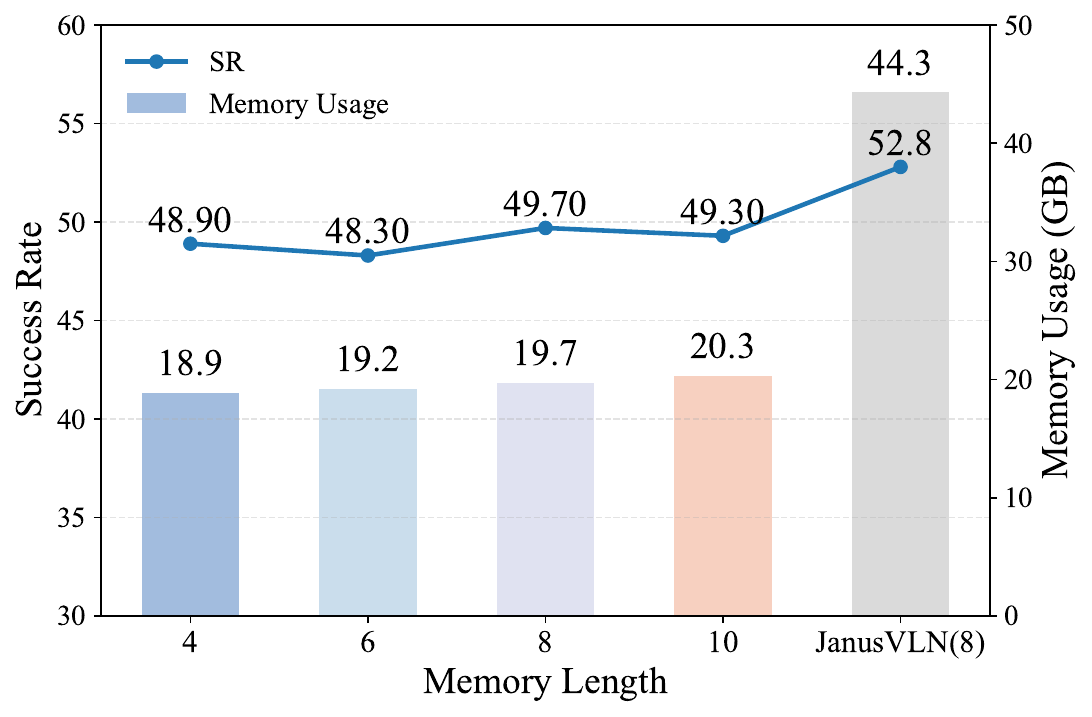}
    \caption{Memory usage and SR comparison in R2R dataset with different memory length}
    \label{fig:fi4}
\end{figure}

We conduct the ablation study on the memory length and efficiency. We conduct an ablation study on the memory length by setting the memory size to 4, 6, 8, and 10, respectively. As shown in Figure 4, with the memory capacity increases, the overall performance of our method improves steadily, without obvious performance fluctuations. This indicates that our method is robust to the choice of memory size.

Combined with the peak GPU memory usage and inference time, we further observe that the memory overhead does not increase as fast as the memory size increases, and the peak GPU memory remains below 24GB under all settings, and the inference time per step is 59ms. These results demonstrate that our method enables practical inference on edge-level devices. In contrast, JanusVLN requires approximately 44.3GB peak GPU memory, and inference time is 194ms, which is 2x than our GPU usage and 4x than time usage with only about 3\% SR decrease, highlighting the clear advantage of our method in inference efficiency and deployment friendliness.

Furthermore, although JanusVLN achieves higher results on some performance metrics, our method provides a more practical, and resource-friendly alternative for VLN. Compared with methods that rely on high-memory servers or complex external modules, our approach shows clear advantages in resource constrained scenarios or on devices with limited GPU memory, making it easier to deploy on edge devices and reducing dependence on high-performance hardware. Therefore, users can flexibly choose the most suitable navigation model according to their specific application requirements, balancing performance, resource consumption, and deployment cost.

\paragraph{Other Ablations} We also conduct experiments about the \textbf{future state modeling analysis}, \textbf{memory management analysis}, \textbf{data ablation}, and \textbf{dataset transfer ability} to show the effectiveness of LookStep, which is described in the Appendix~\ref{append:other}.

\section{Conclusion}

In this paper, we proposed \textbf{LookStep}, an end-to-end VLN framework with resource efficiency. LookStep explicitly models candidate-action future states with LFS and selectively preserves navigation-critical observations through EDRM. We empirically validate the benefits of future-state-based decision making and event-level memory. Experiments on VLN-CE benchmarks show that LookStep achieves stronger navigation performance with lower time and memory overhead, demonstrating the effectiveness of combining future-oriented action evaluation with selective historical memory for VLN.

\section*{Limitations}
Our method has certain limitations. Due to resource constraints, it is trained only on the aforementioned datasets and has not been further enhanced with training paradigms such as DAgger. As a result, there remains a performance gap compared with large-scale methods trained with more extensive data and optimization strategies. In addition, its compatibility with such advanced training paradigms has not yet been fully explored.
\section*{Acknowledgment}
This research was supported by the Jiangsu Science Foundation (BK20243012,BG2024036),Natural Science Foundation of China (62576162, 624B2068), and the Fundamental Research Funds for the Central Universities (022114380023).

\bibliography{custom}


\appendix
\section{Additional Analysis of the Theoretical Motivation}
  \label{app:theory111}

  This appendix provides additional details for the oracle-level motivation
  introduced in Sec.~\ref{sec:theory}. The analysis is intended to clarify the
  information represented by the future-state labels and its relationship to
  the generated-state process, rather than to provide an unconditional
  performance guarantee for the implemented system.

  \subsection{Oracle-Level Interpretation}

  Let $X_t$ denote the navigation context, $a_{t+1}^{\star}$ the expert action,
  and $F_t^{\star}$ the future-state labels constructed from the expert
  trajectory. For any conditioning variable $Z$ and action predictor
  $q(\cdot\mid Z)$, its expected log risk can be decomposed as
  \begin{equation}
  \begin{aligned}
  \mathcal{R}_{\log}(q;Z)
  &=\mathbb{E}\!\left[-\log q(a_{t+1}^{\star}\mid Z)\right]\\
  &=H(a_{t+1}^{\star}\mid Z) \\
  &+\mathbb{E}_{Z}\!\left[
  \mathrm{KL}\!\left(
  P(\cdot\mid Z)\,\|\,q(\cdot\mid Z)
  \right)\right].
  \end{aligned}
  \label{eq:app-log-risk}
  \end{equation}
  Therefore, the minimum is attained by
  $q^{\star}(\cdot\mid Z)=P(\cdot\mid Z)$ and equals
  $H(a_{t+1}^{\star}\mid Z)$.

  For direct action prediction and oracle future-state-conditioned prediction,
  the corresponding Bayes-optimal log risks are
  \begin{equation}
  \mathcal{R}_{\mathrm{base}}^{\star}
  =H(a_{t+1}^{\star}\mid X_t),
\end{equation}
\begin{equation}
  \mathcal{R}_{\mathrm{oracle\text{-}fsm}}^{\star}
  =H(a_{t+1}^{\star}\mid X_t,F_t^{\star}).
  \end{equation}
  By the definition of conditional mutual information,
  \begin{equation}
  \begin{aligned}
  \mathcal{R}_{\mathrm{base}}^{\star}
  -\mathcal{R}_{\mathrm{oracle\text{-}fsm}}^{\star}
  &=H(a_{t+1}^{\star}\mid X_t)\\
  &-H(a_{t+1}^{\star}\mid X_t,F_t^{\star})\\
  &=I(a_{t+1}^{\star};F_t^{\star}\mid X_t)
  \geq 0.
  \end{aligned}
  \label{eq:app-oracle-gap}
  \end{equation}
  The inequality is strict if and only if
  $I(a_{t+1}^{\star};F_t^{\star}\mid X_t)>0$.
  This identity quantifies the action-relevant information represented by the
  oracle future-state labels and motivates making candidate-action outcomes
  explicit before predicting the final action.

  Importantly, Eq.~\eqref{eq:app-oracle-gap} compares predictors with and
  without access to the oracle variable $F_t^{\star}$. It is an idealized motivation.

  \subsection{Generated-State Approximation}

  To characterize the effect of intermediate-generation errors, let
  $h(X_t,F)$ be the deterministic decoded action when the future-state
  representation is $F$. Define
  \begin{equation}
  \begin{aligned}
  \epsilon_{\mathrm{actual\text{-}fsm}}
  &=\Pr[h(X_t,\hat F_t)\neq a_{t+1}^{\star}],\\
  \epsilon_{\mathrm{oracle\text{-}fsm}}
  &=\Pr[h(X_t,F_t^{\star})\neq a_{t+1}^{\star}],\\
  \delta
  &=\Pr[h(X_t,\hat F_t)\neq h(X_t,F_t^{\star})]
  \end{aligned}
  \end{equation}
  For every sample,
  \begin{equation}
  \begin{aligned}
  \mathbf{1}[h(X_t,\hat F_t)\neq a_{t+1}^{\star}]
  \leq&
  \mathbf{1}[h(X_t,F_t^{\star})\neq a_{t+1}^{\star}]\\
  +&\mathbf{1}[h(X_t,\hat F_t)\neq h(X_t,F_t^{\star})]
  \end{aligned}
  \end{equation}
  Taking expectations gives
  \begin{equation}
  \epsilon_{\mathrm{actual\text{-}fsm}}
  \leq
  \epsilon_{\mathrm{oracle\text{-}fsm}}+\delta,
  \delta\leq\Pr[\hat F_t\neq F_t^{\star}]
  \label{eq:app-generated-bound}
  \end{equation}
  Here, $\delta$ measures only intermediate-generation errors that change the
  final action decision. Relative to a learned direct-action predictor with
  error $\epsilon_{\mathrm{base}}$, the condition
  \begin{equation}
  \epsilon_{\mathrm{oracle\text{-}fsm}}+\delta
  <\epsilon_{\mathrm{base}}
  \end{equation}
  is sufficient for the generated-state process to achieve a lower decoded
  action error. This is a characterization of when the oracle benefit survives
  intermediate-generation errors.

  Overall, the oracle identity provides the conceptual motivation for exposing
  action-relevant future states, while the generated-state decomposition
  clarifies the approximation gap in the implemented process. The practical
  effectiveness of the specific LFS formulation is evaluated through the
  experiment results.

\section{Experiment Details}

\label{append:labelset}
\paragraph{Experiment Details}
We constructed our methods based on Qwen3-VL 8B. The model is trained for one epoch using only the R2R-CE and RxR-CE datasets with the ms-swift package and a learning rate of 1e-5. We set the memory length to 8 for the memory queue. We train the model on an Ubuntu Linux Server with 8 NVIDIA A100 80 GB GPUs. The training process is about 1000 GPU hours (PCIe).

The definition of the predefined action set is:
$\{\texttt{MOVE},\texttt{TURN\_LEFT},\texttt{TURN\_RIGHT},\texttt{STOP}\}$

The future-state label set is defined as:
\[
\begin{aligned}
 \{&
\texttt{advance},
\texttt{advance\_after\_turn}, \\
&
\texttt{advance\_to\_goal},
\texttt{start\_turn}, \\
&
\texttt{continue\_turn},
\texttt{finish\_turn},\\
&
\texttt{success\_stop},
\texttt{too\_early}, \\
&
\texttt{overshoot\_goal},
\texttt{wrong\_at\_goal},\\
&
\texttt{premature\_forward},
\texttt{wrong\_turn}, \\
&
\texttt{over\_turn},
\texttt{reverse\_turn},\\
&
\texttt{early\_left\_turn},
\texttt{early\_right\_turn}
\}.
\end{aligned}
\]

The progress label set is:
\[
\{
\texttt{start},
\texttt{early},
\texttt{middle},
\texttt{late},
\texttt{near\_goal}
\}.
\]

The memory-role label set is:
\[
\begin{aligned}
\{&
\texttt{stop\_evidence},
\texttt{start\_view},\\
&
\texttt{turn\_start}, \\
&
\texttt{turn\_end},
\texttt{post\_turn\_alignment},\\
&
\texttt{goal\_approach},
\texttt{recent\_only}
\}.
\end{aligned}
\]

 All labels are automatically generated from expert action trajectories using deterministic rules, without requiring additional manual annotations. First, the \textbf{future-state labels} are constructed according to the current expert action, the
  position of the current step within a consecutive same-direction turning segment, the post-turn action state, and the subsequent $K=5$ expert actions. For the expert action, a forward action is labeled as \texttt{advance\_after\_turn},
  \texttt{advance\_to\_goal}, or \texttt{advance}, depending on whether the current step is the first forward step after a turn or whether \texttt{STOP} occurs within the future window. A turning action is labeled as \texttt{start\_turn},
  \texttt{continue\_turn}, or \texttt{finish\_turn} when it occurs at the beginning, in the interior, or at the end of a consecutive turning segment, respectively; a single-step turn is labeled as \texttt{start\_turn}. An expert \texttt{STOP} action is
  labeled as \texttt{success\_stop}. For candidate actions not executed by the expert, we do not perform additional environment rollouts, but instead construct rule-based counterfactual outcome labels from the expert trajectory. Selecting \texttt{STOP}
  before reaching the goal is labeled as \texttt{too\_early}. Moving forward while the expert is turning is labeled as \texttt{premature\_forward}, while selecting a non-expert turning direction is labeled as \texttt{wrong\_turn}. At the first forward step
  after a turn, continuing to turn in the same direction is labeled as \texttt{over\_turn}, whereas turning in the opposite direction is labeled as \texttt{reverse\_turn}. If a candidate turn matches the first upcoming turn within the future window but is
  executed prematurely, it is labeled as \texttt{early\_left\_turn} or \texttt{early\_right\_turn}. When the expert action is \texttt{STOP}, moving forward and turning are labeled as \texttt{overshoot\_goal} and \texttt{wrong\_at\_goal}, respectively. All
  remaining incorrect candidate actions are labeled as \texttt{wrong\_turn}. Second, the \textbf{progress labels} are generated according to the relative position of the current step in the expert trajectory. The first step is labeled as \texttt{start}; a
  step is labeled as \texttt{near\_goal} if its expert action is \texttt{STOP} or if it is among the final two trajectory steps. The remaining steps are labeled as \texttt{early}, \texttt{middle}, or \texttt{late} when the relative progress $t/(T-1)$ is
  below $0.33$, within $[0.33,0.66)$, or no smaller than $0.66$, respectively. Finally, the \textbf{memory-write and memory-role labels} are constructed using the following rules in priority order. A frame whose expert action is \texttt{STOP} is labeled as
  \texttt{keep}/\texttt{stop\_evidence}, and the first frame of an episode is labeled as \texttt{keep}/\texttt{start\_view}. For a multi-step consecutive same-direction turning segment, its first and last frames are labeled as \texttt{keep}/
  \texttt{turn\_start} and \texttt{keep}/\texttt{turn\_end}, respectively, while its intermediate frames are labeled as \texttt{drop}/\texttt{recent\_only}; a single-step turn is labeled only as \texttt{keep}/\texttt{turn\_start}. The first
  \texttt{MOVE\_FORWARD} frame after a turning segment is labeled as \texttt{keep}/\texttt{post\_turn\_alignment}. When \texttt{STOP} first enters the future window of the subsequent $K=5$ actions, the current frame is labeled as \texttt{keep}/
  \texttt{goal\_approach}. All remaining ordinary frames are labeled as \texttt{drop}/\texttt{recent\_only}.

\paragraph{Compared Methods}

We compare our method with HPN+DN~\citep{krantz2021waypoint}, CMA~\citep{hong2022bridging}, Sim2Sim~\citep{krantz2022sim}, VLN$\circlearrowleft$BERT~\citep{hong2022bridging}, Ego$^2$-Map~\citep{hong2023learning}, DreamWalker~\citep{wang2023dreamwalker}, GridMM~\citep{wang2023gridmm}, Reborn~\citep{wang2023gridmm}, InstructNav~\citep{long2024instructnav}, COSMO~\citep{zhang2025cosmo}, AO-Planner~\citep{chen2025affordances}, LAW~\citep{raychaudhuri2021language}, MapNav~\citep{zhang2025mapnav}, g3D-LF~\citep{wang2025g3d}, Seq2Seq~\citep{krantz2020beyond}, NaVid-4D~\citep{liu2025vid}, NavMorph~\citep{yao2025navmorph}, NaVid~\citep{zhang2024navid}, Sim2Real~\citep{wang2024sim}, StreamVLN~\citep{wei2025streamvln}, Uni-NaVid~\citep{zhang2024uni}, NaVILA~\citep{cheng2024navila}, JanusVLN~\citep{zeng2025janusvln}.

\paragraph{Real world quantitative results.}

In real-world experiments, we use the zsibot-L1 as the robotic platform, equipped with an FPV camera to capture the front RGB. LookStep runs on an RTX 4090 GPU 24G to process the instructions.

We conduct real-world experiments in a challenging workspace environment. This scene contains multiple visually similar and easily confusable objects and involves complex navigation instructions such as entering and exiting rooms. We design 10 instructions, and each is repeated three times. A trial is considered successful if the robot stops within
1 meter of the target. LookStep successfully completes difficult navigation tasks in such a complex scenario with a 70\% success rate.

\section{Other Ablation}
\label{append:other}
\paragraph{Future State Modeling Analysis}

To investigate whether our FSM module can understand the consequence of each candidate action, we further analyze the outcomes predicted by the model. In this analysis, we mainly focus on two critical navigation states: approaching the goal and making turns.

\begin{table}[ht]
\centering
\caption{Success rate after \texttt{advance\_to\_goal} in R2R dataset.}
\label{tab:advance_to_goal_stop}
\begin{tabular}{lc}
\toprule
\textbf{Metric} & \textbf{SR Rate} \\
\midrule
Successful \texttt{STOP}
& $55.52\%$ \\
No \texttt{STOP}
& $0.028\%$ \\
\bottomrule
\end{tabular}
\end{table}

First, we analyze the goal-approaching state. When the model predicts the outcome \texttt{advance\_to\_goal}, it achieves a relatively high success rate in subsequently issuing \texttt{STOP} and completing navigation. In contrast, only one case eventually fails to stop. This indicates that \texttt{advance\_to\_goal} is not produced arbitrarily; instead, the predicted goal-approaching outcome carries meaningful navigation information.

For turning behaviors, when the instruction requires a turn, the model predicts negative outcomes for the opposite action in $81.17\%$ of the cases. This suggests that the predicted outcomes are also instruction-aware and outcome provides an explicit action-evaluation signal: the model can associate candidate actions with their corresponding consequences and learn meaningful action-outcome relationships.

\paragraph{Memory Management Analysis}

To verify whether our method can identify key navigation events and incorporate them into historical observations, we conduct an analysis of the predicted memory roles. As shown in Table~\ref{tab:memory_role_accuracy}, the model achieves high accuracy on most memory roles related to explicit navigation events.

\begin{table}[ht]
\centering
\caption{Accuracy of different memory roles in R2R dataset.}
\label{tab:memory_role_accuracy}
\begin{tabular}{lc}
\toprule
\textbf{Memory Role} & \textbf{Accuracy} \\
\midrule
\texttt{start\_view} & $100\%$ \\
\texttt{post\_turn\_alignment} & $100\%$ \\
\texttt{turn\_start} & $99.51\%$ \\
\texttt{turn\_end} & $98.99\%$ \\
\texttt{stop\_evidence} & $100\%$ \\
\texttt{goal\_approach} & $43.64\%$ \\
\bottomrule
\end{tabular}
\end{table}

These results indicate that the model can reliably identify memory tags tied to explicit action structures. In particular, it accurately recognizes key navigation events such as the start of a turn, the end of a turn, post-turn alignment, and stop evidence. In contrast, goal-approaching keyframes remain more challenging, as they require joint reasoning over the target semantics and the agent's current spatial position.

This observation is further supported by our turning statistics: after \texttt{turn\_end} appears, $99.72\%$ of subsequent actions no longer involve turning. This suggests that the model indeed learns the semantic meaning of ``turn end'': once a frame is tagged as \texttt{turn\_end}, the agent typically transitions to moving forward rather than continuing to turn. Overall, the model can reliably perform event-driven rolling memory writing by selecting keyframes that are important for long-horizon navigation. The role of memory tags is therefore to ensure high-quality historical compression, rather than to independently determine navigation success, which is consistent with our ablation results.

\paragraph{Memory-Write and Cross-Episode Consistency.}
To complement the per-role results above, we analyze the complete R2R Val-Unseen traces, covering 1,839 episodes and 154,481 navigation steps. Since an online agent trajectory may deviate from its expert trajectory, expert memory-role labels cannot be aligned with every executed step. We therefore reconstruct a rule-consistency proxy from each executed action sequence using the same deterministic EDRM rules used to construct the training labels ($K=5$). Consequently, the statistics in Table~\ref{tab:edrm_consistency} measure whether the generated memory decisions consistently follow our predefined operational event rules, rather than universal semantic-event recognition accuracy.

The keep/drop distribution shows that EDRM does not collapse to always retaining or always discarding observations. Its $99.32\%$ keep-decision F1, together with the episode-level and paired-boundary results, indicates that the model applies the predefined event-writing rules consistently across unseen episodes. In particular, both boundaries are correctly identified in $98.26\%$ of evaluable multi-step turn segments, rather than the model merely producing isolated high-frequency role labels. These results support the consistency of the task-specific EDRM decision mechanism, but do not imply that rule consistency alone guarantees navigation success or generalizes to unrestricted open-world event recognition.

\paragraph{Data Ablation}

To evaluate the impact of training data on model performance, we conduct a data ablation study, as shown in Table~\ref{tab:data_ablation}. Compared with using R2R alone, jointly training on R2R and RxR consistently improves performance across all metrics. Specifically, NE decreases from $6.61$ to $5.34$, while SR increases from $40.3\%$ to $49.7\%$ and SPL improves from $36.5\%$ to $45.3\%$. These results indicate that richer instruction and trajectory data can enhance the model's understanding of navigation semantics, path structures, and long-horizon decision making. They also demonstrate that our method can effectively leverage additional training data and translate it into stable performance gains.

\begin{table}[ht]
\centering
\caption{Data ablation results in R2R dataset}
\label{tab:data_ablation}
\begin{tabular}{l|cccc}
\toprule
Method & NE$\downarrow$ & OS$\uparrow$ & SR$\uparrow$ & SPL$\uparrow$ \\
\midrule
R2R+RxR & 5.34 & 55.9 & 49.7 & 45.3 \\
R2R & 6.61 & 48.3 & 40.3 & 36.5 \\
\bottomrule
\end{tabular}
\end{table}
\paragraph{Dataset Transfer Ability}

To further verify the transferability of our method across different datasets and task settings, we conduct experiments on the HM3D-OVON~\citep{yokoyama2024hm3d} val-unseen split. As shown in Table~\ref{tab:hm3d_ovon_val_unseen}, our method achieves strong performance, reaching an SR of $38.0\%$ and an SPL of $26.9\%$. The results show that our method not only improves the task completion rate but also generates more efficient navigation trajectories. This indicates that our method generalizes well to different datasets and task settings, demonstrating strong cross-dataset transfer ability~\citep{yokoyama2024vlfm}.

\begin{table}[ht]
\centering
\caption{Comparison on a HM3D-OVON val unseen subset.}
\label{tab:hm3d_ovon_val_unseen}
\begin{tabular}{l|cc}
\toprule
\textbf{Method} & \textbf{SR$\uparrow$} & \textbf{SPL$\uparrow$} \\
\midrule
VLFM  & 35.2 & 19.6 \\
DAgRL+OD & 37.1 & 19.8 \\
Ours & \textbf{38.0} & \textbf{26.9} \\

\bottomrule
\end{tabular}
\end{table}

\paragraph{Controlled Ablations.}
We provide a controlled ablation results.
\begin{table}[h]
\centering
\small
\setlength{\tabcolsep}{3.5pt}
\caption{Controlled ablations.}
\label{tab:controlled_lfs_ablation}
\begin{tabular}{lcccc}
\toprule
\textbf{Method} & \textbf{NE}$\downarrow$ & \textbf{OS}$\uparrow$ & \textbf{SR}$\uparrow$ & \textbf{SPL}$\uparrow$ \\
\midrule
LookStep & \textbf{5.34} & \textbf{55.9} & \textbf{49.7} & \textbf{45.3} \\
LookStep w/o LFS & 5.39 & 52.8 & 46.9 & 42.4 \\
LookStep w/o EDRM* & 6.64 & 70.9 & 37.4 & 22.1 \\
LookStep w/o EDRM & 5.53 & 51.7 & 45.2 & 42.7 \\
Action-only IL & 6.01 & 50.1 & 41.7 & 39.5 \\
LookStep w/o \texttt{outcome} & 5.37 & 53.2 & 47.8 & 43.9 \\
LookStep w/o \texttt{progress} & 5.35 & 53.7 & 48.3 & 43.6 \\
\bottomrule
\end{tabular}
\end{table}
As shown in Table~\ref{tab:controlled_lfs_ablation}, the results show that LookStep achieves a best performance in SR, emoving any component leads to a drop in success rate, indicating that both LFS and EDRM are indispensable.

\begin{table*}[t]
\centering
\small
\caption{Additional EDRM write-decision and cross-episode consistency statistics on R2R Val-Unseen.}
\label{tab:edrm_consistency}
\begin{tabular}{llc}
\toprule
\textbf{Analysis} & \textbf{Metric} & \textbf{Result} \\
\midrule
Coverage & Episodes / steps & 1,839 / 154,481 \\
Memory write & Predicted keep / drop frequency & $53.93\%$ / $46.07\%$ \\
Memory write & Keep precision / recall / F1 & $99.37\%$ / $99.28\%$ / $99.32\%$ \\
Cross-episode & Mean / median keep-F1 & $98.79\%$ / $100.00\%$ \\
Cross-episode & Episodes with keep-F1 $\geq 95\%$ & $93.31\%$ \\
Turn segment & Multi-step turn segments & 9,746 \\
Turn segment & Both start/end boundaries correctly identified & $98.26\%$ \\
Turn segment & Interior steps labeled \texttt{drop}/\texttt{recent\_only} & $99.98\%$ \\
Turn segment & Episodes with all turn pairs correctly identified & $94.23\%$ \\
\bottomrule
\end{tabular}
\end{table*}
\section{Failure Case Analysis}

\begin{figure*}[ht]
    \centering
    \includegraphics[width=\linewidth]{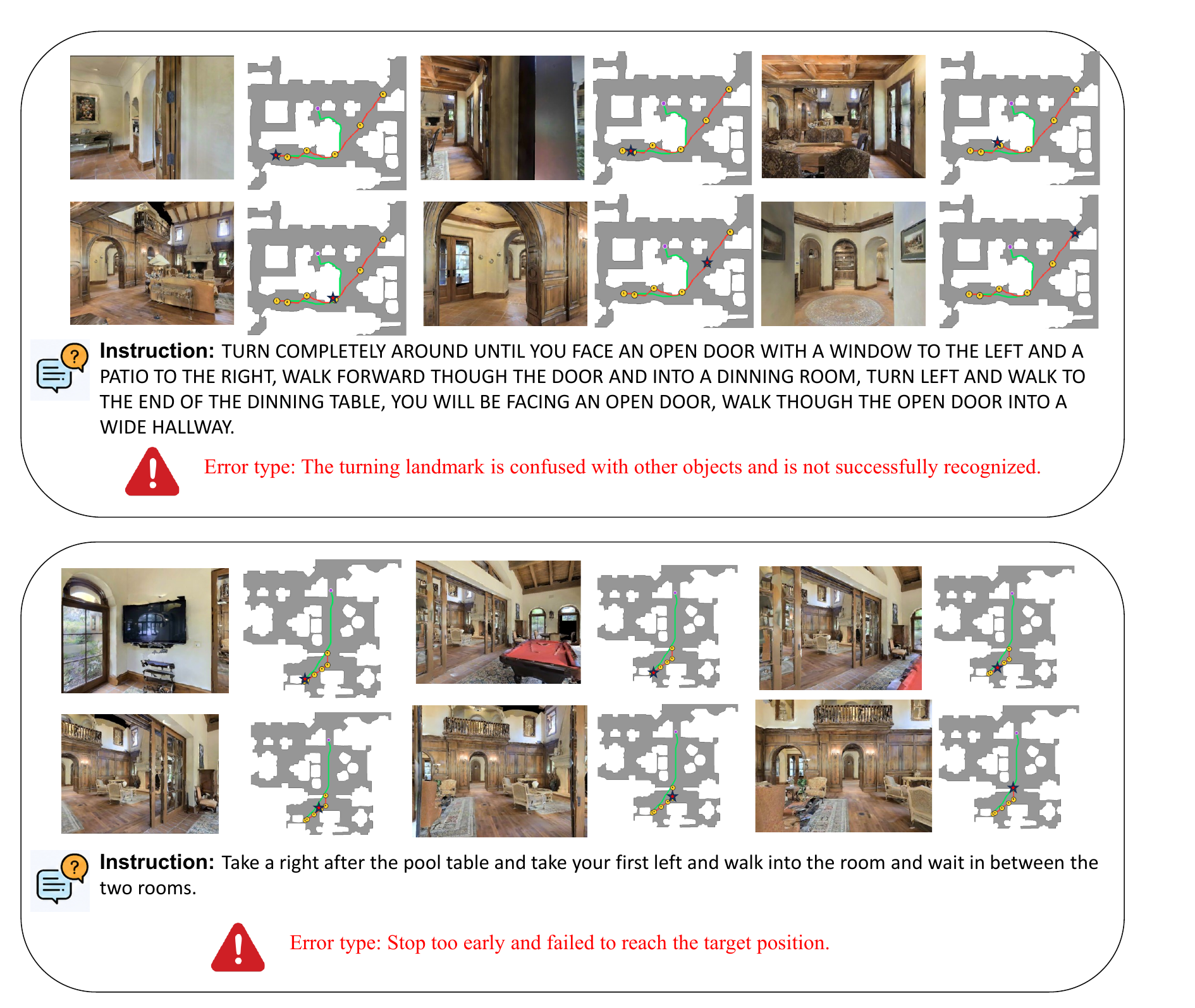}
    \caption{Visualization and presentation of the types of failure cases in R2R dataset.}
    \label{fig:fail}
\end{figure*}

We identify two representative failure modes of LookStep and visualize them in Figure~\ref{fig:fail}. As shown in the figure, LookStep usually follows the optimal trajectory direction at the beginning of these failure cases. However, when multiple visually similar objects appear in the scene, the model may confuse the turning landmark with other objects, leading to failed landmark recognition. In addition, LookStep occasionally misidentifies the correct stopping position and stops outside the success region, resulting in navigation failure. We conjecture that this issue may stem from the limited alignment between the MLLM's scale estimation and real-world spatial scale, while fine-grained object recognition in complex scenes remains challenging.

\section{More Qualitative Results}
\begin{figure*}
    \centering
    \includegraphics[width=\linewidth]{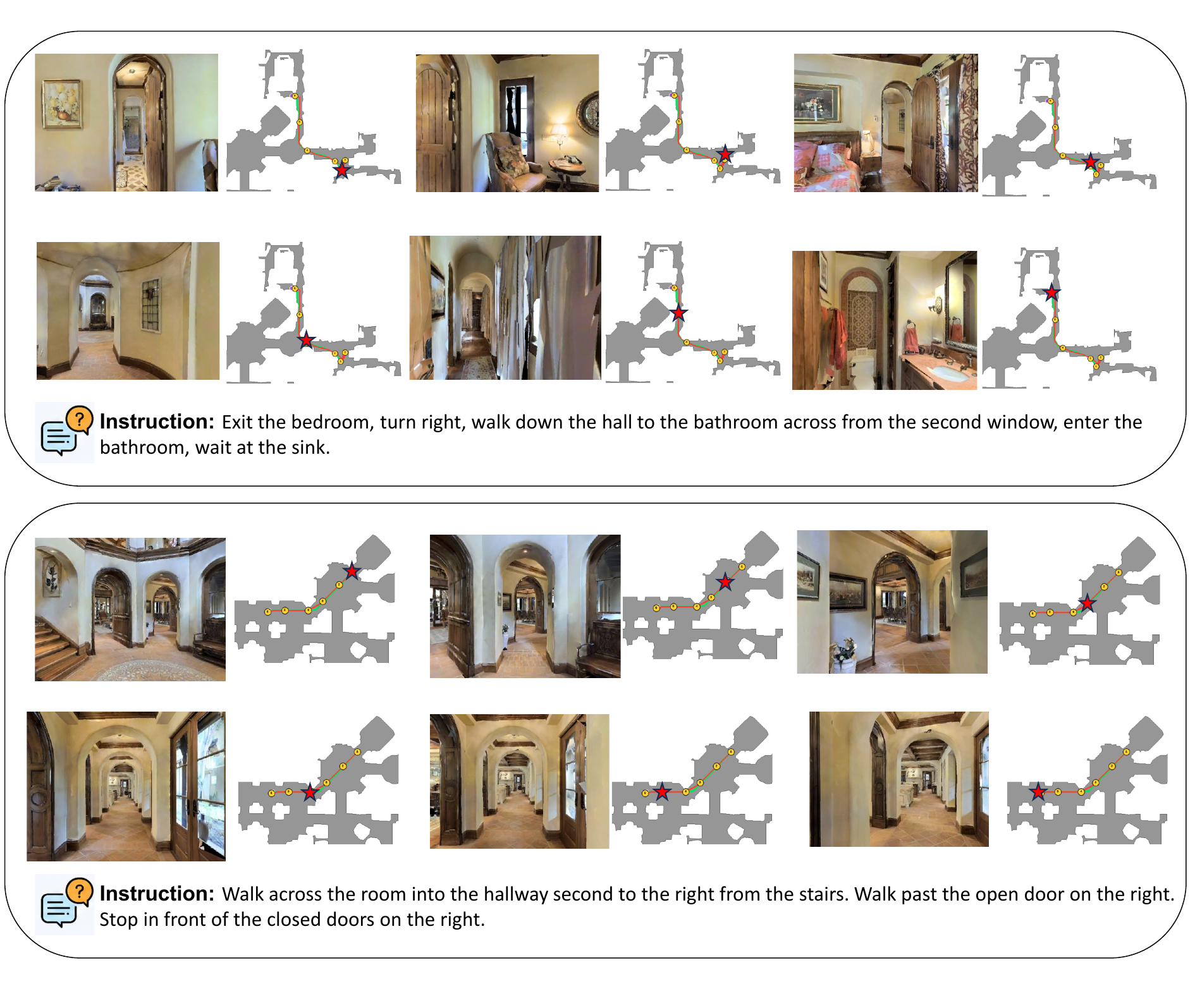}
    \caption{Visualization of LookStep on the R2R benchmark.}
    \label{fig:vlnce_visualization}
\end{figure*}
\begin{figure*}
    \centering
    \includegraphics[width=\linewidth]{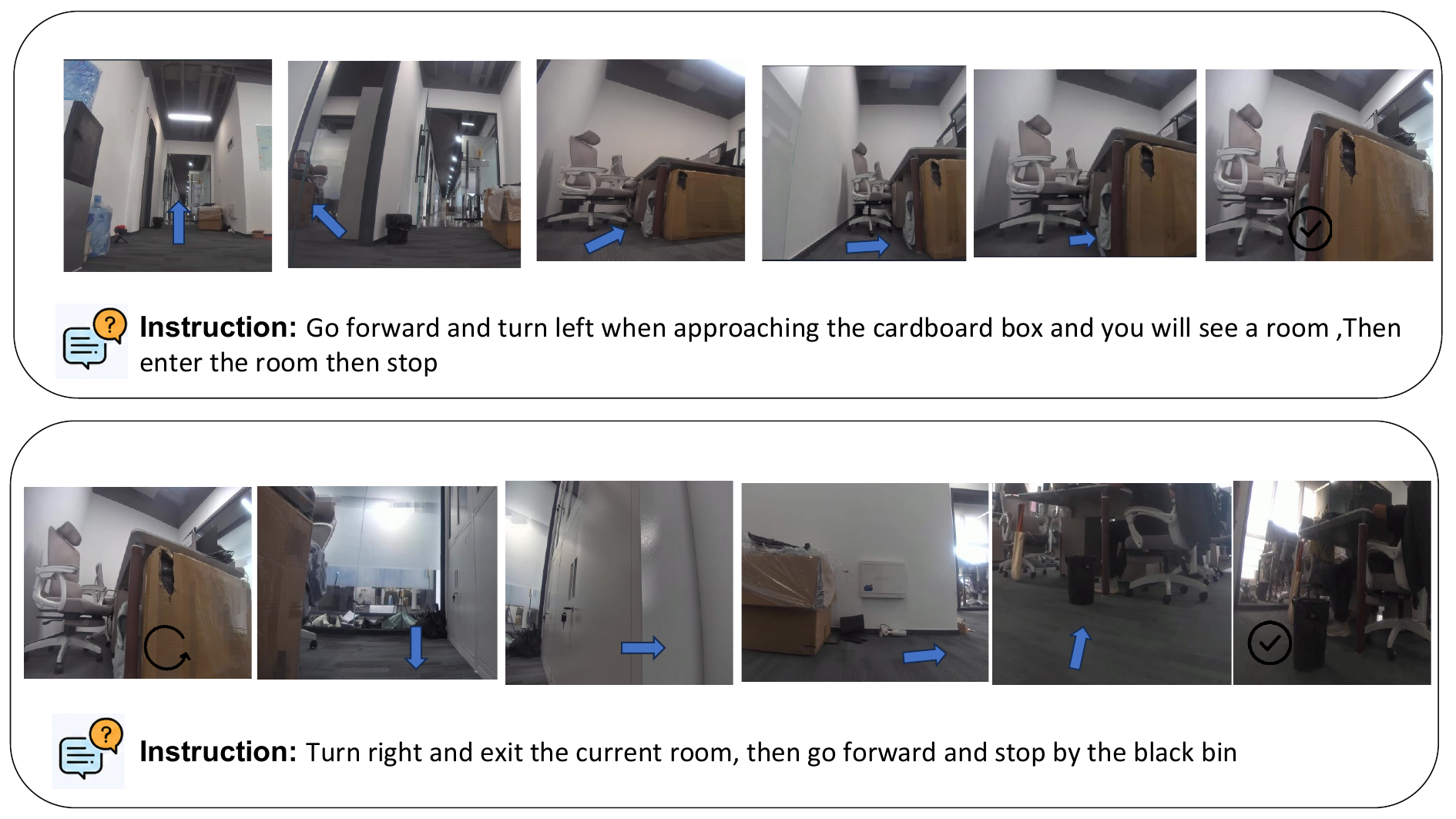}
    \caption{Visualization of the first-person perspective in real-world experiments.}
    \label{fig:realworld_first_person}
\end{figure*}
\paragraph{Visualization of LookStep in VLN-CE Benchmarks}
To further demonstrate the practical behavior of LookStep in different navigation scenarios, we provide additional qualitative visualizations. First, we show the navigation process of LookStep on the VLN-CE benchmark. As shown in Figure~\ref{fig:vlnce_visualization}, LookStep can progressively understand the current state according to the language instruction and make reasonable forward, turning, and stopping decisions at key locations. These visualizations indicate that LookStep can not only complete navigation tasks in standard simulated environments, but also maintain strong instruction-following and spatial decision-making abilities over long-horizon trajectories.

\paragraph{First-Person Perspective in Real-World Experiments}
In addition, we present first-person perspective results from real-world experiments. As shown in Figure~\ref{fig:realworld_first_person}, LookStep can make stable navigation decisions based on current observations in real-world scenes, while recognizing key landmarks, completing turns, and approaching the target in complex environments. This further demonstrates that, although LookStep is mainly trained on simulated data, it still exhibits strong real-world generalization ability.

\section{Usage of LLMs}
In this work, we used large language models (LLMs) only in a limited, supportive capacity. All key research ideas, theoretical analyses, experimental designs, and the writing of the main text were carried out independently by the authors. LLMs were not used to generate or edit any scientific content in the manuscript, nor did they contribute to the formulation of research hypotheses or the interpretation of results. The authors take full responsibility for the accuracy, originality, and completeness of all content in the paper. The use of LLM is only about aiding or polishing writing in this paper.

\label{sec:appendix}

\end{document}